\documentclass[letterpaper, 10pt, journal, twoside]{IEEEtran}
\usepackage[utf8]{inputenc}
\usepackage[english]{babel}

\usepackage[lined,ruled]{algorithm2e}
\usepackage{amsmath, amssymb, amsfonts}
\usepackage{bm}
\usepackage{booktabs}
\usepackage{cite}
\usepackage{color}
\usepackage{graphicx}
\usepackage{siunitx}
\usepackage{subcaption}
\usepackage{url}
\usepackage{hyperref}
\usepackage{balance}

\newcommand{\thetitle}{Vision-based Drone Flocking in Outdoor Environments}

\graphicspath{{./figures/}}

\title{\huge\thetitle}
\author{
Fabian Schilling, Fabrizio Schiano, and Dario Floreano%
\thanks{Manuscript received: December 10, 2020; accepted February 2, 2021. This letter was recommended for publication by Editor P. Pounds upon evaluation of the reviewer comments.
This work was supported by the Swiss National Science Foundation with grant number 200021-155907 (\textit{Corresponding author: Fabian Schilling}).}
\thanks{The authors are with the Laboratory of Intelligent Systems, École Polytechnique Fédérale de Lausanne, 1015 Lausanne, Switzerland (e-mail: \href{mailto:fabian.schilling@epfl.ch}{fabian.schilling@epfl.ch}; \href{mailto:fabrizio.schiano@epfl.ch}{fabrizio.schiano@epfl.ch}; \href{mailto:dario.floreano@epfl.ch}{dario.floreano@epfl.ch}).}
}

\newcommand{\tit}[1]{\textit{#1}}

\newcommand{\mbf}[1]{\mathbf{#1}}
\newcommand{\mcl}[1]{\mathcal{#1}}

\newcommand{\txt}[1]{\text{#1}}

\newcommand{\vel}{v}
\newcommand{\pos}{p}
\newcommand{\rel}{r}
\newcommand{\velvec}{\mbf{\vel}}

\newcommand{\relvec}{\mbf{\rel}}

\newcommand{\norm}[1]{\|#1\|}

\newcommand{\set}[1]{\mcl{#1}}

\newcommand{\neighbors}{\set{A}}

\newcommand{\lr}{\eta}
\newcommand{\wdecay}{\lambda}
\newcommand{\loss}{\mcl{L}}
\newcommand{\momentum}{\mu}

\newcommand{\state}{\mbf{x}}
\newcommand{\obs}{\mbf{z}}
\newcommand{\normal}{\mcl{N}}
\newcommand{\mean}{\mbf{m}}
\newcommand{\cov}{\mbf{P}}
\newcommand{\weight}{w}
\newcommand{\controlinput}{\mbf{u}}
\newcommand{\process}{v}
\newcommand{\dt}{\Delta}
\newcommand{\range}{d}
\newcommand{\bearing}{\beta}
\newcommand{\std}{\sigma}
\newcommand{\zeros}{\mbf{0}}
\newcommand{\ones}{\mbf{I}}
\newcommand{\birth}{\gamma}
\newcommand{\clutter}{\kappa}
\newcommand{\prob}{p}
\newcommand{\ii}{{(i)}}
\newcommand{\jj}{{(j)}}
\newcommand{\Q}{\mbf{Q}}
\newcommand{\F}{\mbf{F}}
\newcommand{\B}{\mbf{B}}
\newcommand{\HH}{\mbf{H}}
\newcommand{\K}{\mbf{K}}
\newcommand{\R}{\mbf{R}}

\DeclareMathOperator{\atantwo}{atan2}
\DeclareMathOperator{\diag}{diag}

\begin{document}

\maketitle


\begin{abstract}
Decentralized deployment of drone swarms usually relies on inter-agent communication or visual markers that are mounted on the vehicles to simplify their mutual detection.
This letter proposes a vision-based detection and tracking algorithm that enables groups of drones to navigate without communication or visual markers.
We employ a convolutional neural network to detect and localize nearby agents onboard the quadcopters in real-time.
Rather than manually labeling a dataset, we automatically annotate images to train the neural network using background subtraction by systematically flying a quadcopter in front of a static camera.
We use a multi-agent state tracker to estimate the relative positions and velocities of nearby agents, which are subsequently fed to a flocking algorithm for high-level control.
The drones are equipped with multiple cameras to provide omnidirectional visual inputs.
The camera setup ensures the safety of the flock by avoiding blind spots regardless of the agent configuration.
We evaluate the approach with a group of three real quadcopters that are controlled using the proposed vision-based flocking algorithm.
The results show that the drones can safely navigate in an outdoor environment despite substantial background clutter and difficult lighting conditions.
The source code, image dataset, and trained detection model are available at \href{https://github.com/lis-epfl/vswarm}{https://github.com/lis-epfl/vswarm}.
\end{abstract}

\begin{IEEEkeywords}
Aerial Systems: Perception and Autonomy, Multi-Robot Systems, Sensor-based Control
\end{IEEEkeywords}





\section{Introduction}\label{sec:introduction}

\IEEEPARstart{D}{rone swarms} have a large socio-economic potential and can serve in a variety of real-world applications \cite{floreano_science_2015}.
For example, drones can be leveraged to automatically monitor crops, safely inspect confined spaces, or quickly deliver medicine to inaccessible locations.
Operating these vehicles in swarms could bring increased robustness to failures, larger area coverage, and faster task completion times \cite{coppola_survey_2020}.

Despite this potential, decentralized control has been a limiting factor in the deployment of drone swarms.
For instance, large groups of quadcopters can be used to perform awe-inspiring aerial choreographies in the night sky.
These robotic light shows are a true feat of engineering but individual drones are far from autonomous: their motion is centrally controlled by a ground computer that precomputes their trajectories and continuously monitors their positions.
Hence, the ground computer represents a single point of failure.
Researchers attempted to remove the central computer by equipping drones with hardware that allows them to wirelessly communicate with each other.
Notable examples of these decentralized swarms feature bearing-only formation control \cite{schiano_bearing_2017}, exploration of unknown environments \cite{mcguire_minimal_2019}, as well as flocking with ten fixed-wing drones \cite{hauert_reynolds_2011} or thirty quadcopters \cite{vasarhelyi_optimized_2018}.

\begin{figure}[t]
    \centering
    \includegraphics[width=\columnwidth]{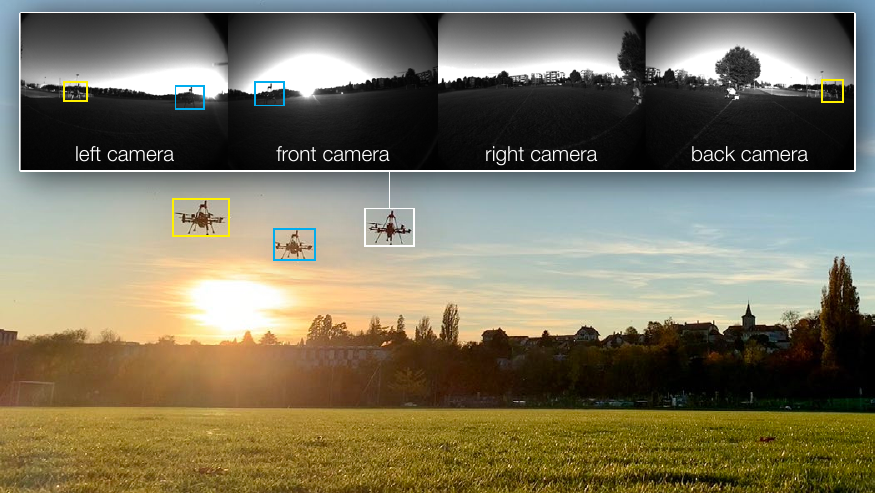}
    \caption{%
        Photo taken during outdoor experiments with detection annotations.
        The quadcopters avoid collisions with each other and remain cohesive as a group while performing a variety of navigation tasks.
        Each agent detects its neighbors in real-time from omnidirectional images.
        There is no communication of state information between agents, and relative positions and velocities are estimated onboard using local visual inputs.
    }\label{fig:overview}
\end{figure}

\newcommand{\mwidth}{0.32 \textwidth}
\begin{figure*}[t]
    \begin{subfigure}{\textwidth}
        \includegraphics[width=\textwidth]{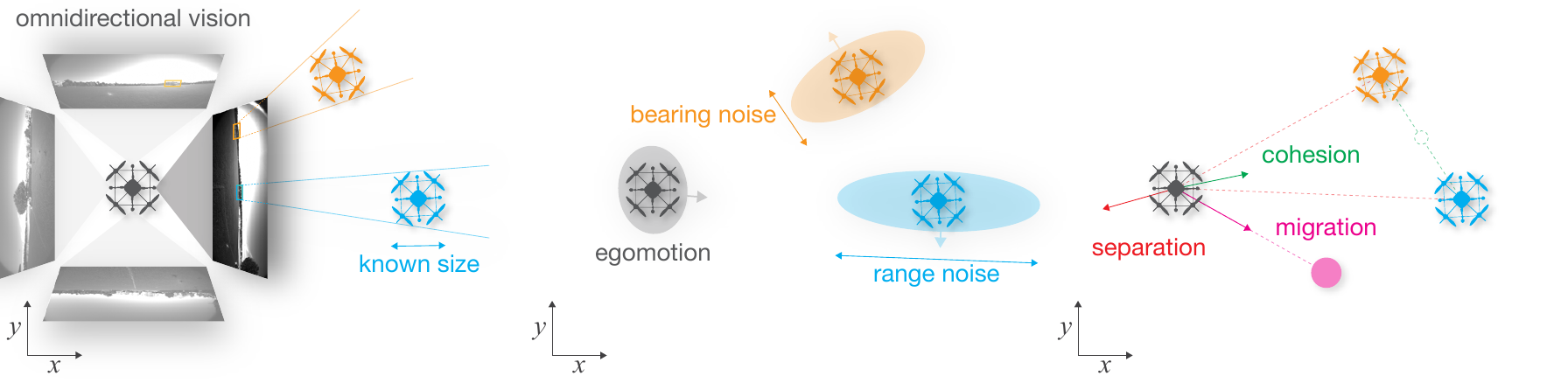}
    \end{subfigure}
    \medskip
    \\
    \begin{subfigure}{\mwidth}
        \caption{Real-time drone detection}\label{fig:method_detection}
    \end{subfigure}
    \begin{subfigure}{\mwidth}
        \caption{Multi-agent state tracking}\label{fig:method_tracking}
    \end{subfigure}
    \begin{subfigure}{\mwidth}
        \caption{Potential-field-based control}\label{fig:method_control}
    \end{subfigure}
    \caption{Overview of the processing steps of the vision-based flocking algorithm for a single time step: a) \textit{detection}, b) \textit{tracking}, and c) \textit{control}.
    Step 1 (\textit{detection}): we detect neighboring agents from omnidirectional images to estimate their relative range and bearing from the camera intrinsics and the drone's known physical size.
    Step 2 (\textit{tracking}): we track the positions and velocities of the detected agents using a linearized observation model of range and bearing, as well as a state transition model that takes the focal agent's egomotion into account.
    Step 3 (\textit{control}): we control the focal agent using a Reynolds-rules-based flocking algorithm that keeps the swarm collision-free and cohesive while following a navigation goal.
    }\label{fig:method}
\end{figure*}

Scaling up decentralized drone swarms is complicated by the limitations of wireless communication.
As the number of robots increases, the communication channels may become saturated and possibly jammed since the data transfer volume scales quadratically with the robot count \cite{cieslewski_data-efficient_2018}.
Frequent retransmissions of messages can lead to delays that render the control of each drone extremely difficult.
Researchers have experimented with different sensory modalities such as sound \cite{basiri_onboard_2016}, but vision seems to be the most scalable approach to address the relative localization problem.
Indeed, there is evidence to support that vision is the primary sensory modality that enables collective motion in animal groups \cite{strandburg-peshkin_visual_2013}.

However, visual detection of outdoor flying drones is challenging because they are relatively small and can fly in environments with large amounts of background clutter and difficult lighting conditions \cite{rozantsev_detecting_2017}.
Additionally, the visual detection algorithm must run onboard the drones, whose own motion may generate motion blur and whose small dimensions and lightweight mass can restrict computational capabilities.
To simplify the relative localization problem, researchers have mounted different types of easily detectable visual markers on the drones \cite{roelofsen_reciprocal_2015,walter_uvdar_2019}.
However, this approach is less general since specialized hardware has to be mounted on each drone, which can result in increased weight and drag, thus reducing the energetic autonomy of the drones.

In recent years, markerless detection of drones has become an active research topic.
In \cite{sapkota_vision-based_2016}, the authors use a boosted cascade of classifiers in combination with visual tracking and finite set filtering to estimate the positions and velocities of markerless drones from a mostly static observer.
However, the method is applied in post-processing and not validated in a sense-and-avoid setting.
In \cite{vrba_marker-less_2020}, the authors propose a combination of stereo vision and convolutional detection to enable leader-follower flight.
However, the to-be-localized agent is always visible in front of the clear sky which simplifies the detection problem and would be impossible to guarantee in a self-organized flock.
Moreover, the limited field of view and processing delays in the proposed system are known to be problematic when flying in dense swarms \cite{vasarhelyi_optimized_2018,soria_influence_2019}.
Other notable examples of markerless detection include approaches based on template matching and morphological filtering \cite{opromolla_vision-based_2018}, as well as convolutional neural networks \cite{wyder_autonomous_2019}.

Our previous work \cite{schilling_learning_2019} proposes a fundamentally different approach to visual flocking based on imitation learning.
Rather than detecting neighboring agents, we predict flocking algorithm commands directly from omnidirectional visual inputs which allow the drones to remain collision-free and cohesive.
The approach is validated with leader-follower experiments in an indoor environment but its reliance on end-to-end learning means that the entire monolithic neural network has to be retrained each time the task and/or visual appearance of the drones change.
Adopting a more modular approach can be beneficial since the flocking algorithm may easily be exchanged for another task-dependent controller, and only the detector would have to be retrained for different drone appearances or environmental conditions.
In all of the above cases, the algorithms are validated on a single agent and not in a multi-robot control setting.

Here, we propose a modular detection and tracking algorithm that enables collision-free and cohesive navigation for drone swarms.
We automatically label images of drones using background subtraction to generate a dataset for the drone detector.
We show that the detector can localize other drones in the presence of background clutter from onboard a flying drone despite being trained on images from a static camera.
We assess the method with a dense group of three real quadcopters that flock in planar configurations in an outdoor environment with difficult lighting conditions.
The omnidirectional camera configuration of the drone is specifically designed to enable safe operation in swarms regardless of the agent configuration.
The overall proposed flocking algorithm is modular since each component, i.e. detection, tracking, and control, is self-contained and can be evaluated independently.
To the best of our knowledge, this is the first entirely vision-based flock that does not depend on visual markers to simplify mutual detections.


\section{Method}\label{sec:method}

The proposed approach to vision-based flocking can be divided into the following steps: \textit{detection}, \textit{tracking}, and \textit{control} (Fig.~\ref{fig:method}).
Firstly, the \textit{detection} module (Fig.~\ref{fig:method_detection}) takes grayscale images from an omnidirectional camera setup as inputs and outputs bounding boxes of nearby drones in real-time (Sec.~\ref{sec:control}).
Secondly, the \textit{tracking} module (Fig.~\ref{fig:method_tracking}) transforms the bounding boxes into range and bearing measurements using the known dimensions of the drones.
Their relative positions and velocities are then estimated from the noisy measurements using a multi-agent state tracker (Sec.~\ref{sec:localization}).
Finally, the \textit{control} module (Fig.~\ref{fig:method_control}) applies a flocking algorithm to the relative positions to obtain high-level control commands that keep the drones collision-free and cohesive (Sec.~\ref{sec:control}).
In the tracking and control steps, we assume that the agents are moving on a horizontal plane.
We mainly introduce this constraint to be able to attribute their mutual repulsion to the flocking algorithm and avoid the effects of physical downwash that may be caused by nearby agents.

\subsection{Real-time monocular drone detection}\label{sec:detection}

\subsubsection{Automatic drone labeling with background subtraction}\label{sec:background_subtraction}

\begin{figure}[t]
    \centering
    \includegraphics[width=\columnwidth]{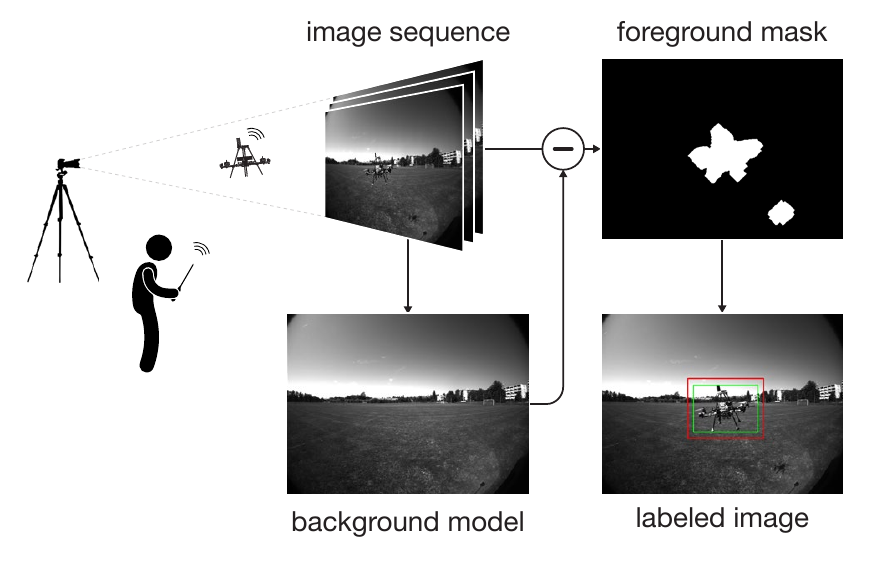}
    \caption{%
        Schematic overview of the automatic dataset generation process.
        The foreground mask is generated using background subtraction.
        We find that the most precise bounding box labels are obtained by dilating the foreground mask since it eliminates discontinuities that occur due to the mechanical design of the drone.
        Enclosing the dilated mask with a bounding box (red rectangle) overestimates the size of the drone.
        We therefore scale the bounding box down (green rectangle) to obtain a precise label.
        We record six flights of roughly one-minute duration in the above target environment.
    }\label{fig:background_subtraction}
\end{figure}

We use background subtraction to automatically generate a labeled image dataset for the object detector (Fig.~\ref{fig:background_subtraction}).
We record images from a stationary camera mounted on a tripod and manually fly a quadcopter within its field of view.
The image data is recorded under varying lighting conditions in both indoor and outdoor environments that contain large amounts of background clutter (Fig.~\ref{fig:dataset}).
For each scene, the camera location and orientation are carefully chosen such that the quadcopter is the dominant source of motion.
The final dataset consists of $9\,891$ training and $1\,931$ testing examples.

In post-processing, we apply a nearest neighbor background subtraction algorithm \cite{zivkovic_efficient_2006} to the sequence of images to learn a background model of the scene.
We extract a foreground mask of the moving parts of the image (and therefore the quadcopter) by computing the element-wise difference of the input image and the background, followed by a thresholding step.
The ground truth label is obtained by filtering out the largest contour present in the foreground mask and enclosing it with an axis-aligned rectangular bounding box.

\newcommand{\datawidth}{0.49 \columnwidth}

\begin{figure}[t]
    \begin{subfigure}{\datawidth}
        \includegraphics[width=\textwidth]{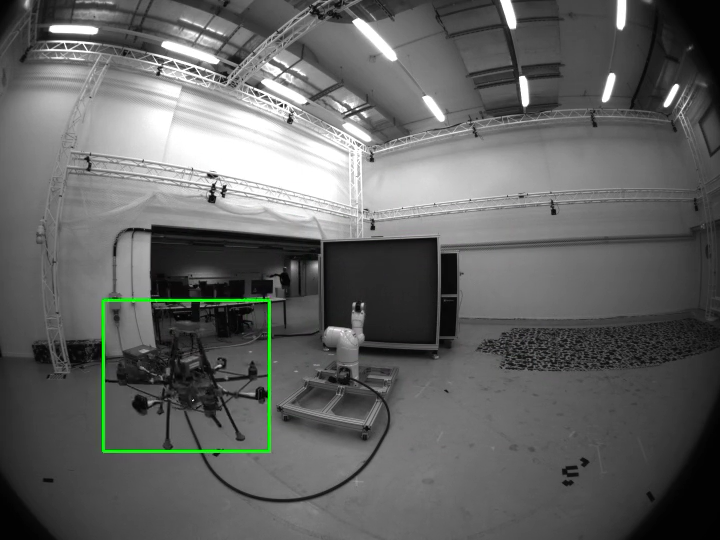}
    \end{subfigure}
    \hfill
    \begin{subfigure}{\datawidth}
        \includegraphics[width=\textwidth]{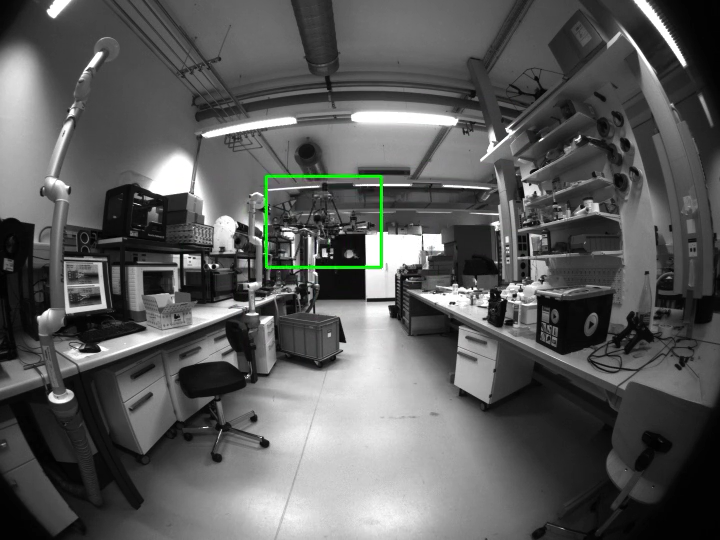}
    \end{subfigure}

    \medskip

    \begin{subfigure}{\datawidth}
        \includegraphics[width=\textwidth]{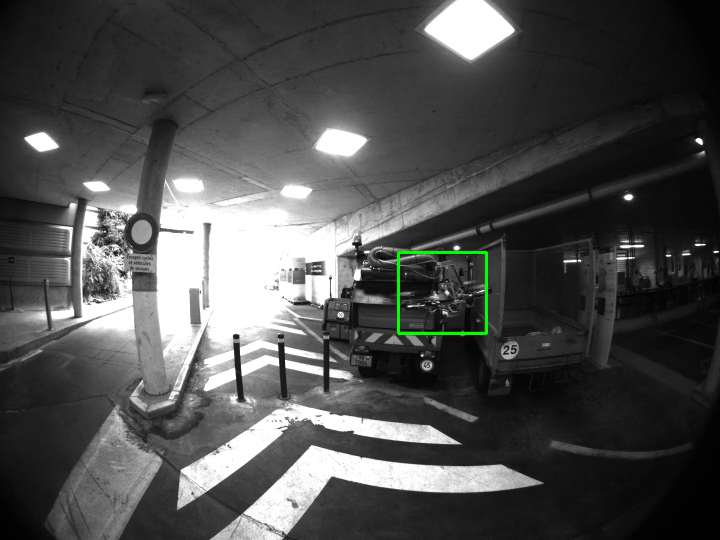}
    \end{subfigure}
    \hfill
    \begin{subfigure}{\datawidth}
        \includegraphics[width=\textwidth]{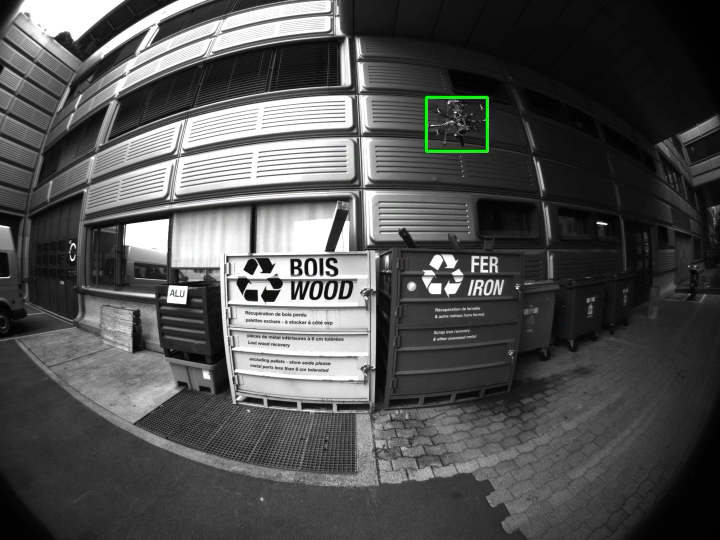}
    \end{subfigure}
    \caption{%
        Example images with bounding box annotations from the dataset generated using background subtraction.
        We record image data from a static camera in both indoor (top row) and outdoor (bottom row) environments.
        The environments are selected to maximize the variety of background clutter and lighting conditions.
        We record three flights of roughly one-minute duration in each of the above environments.
    }\label{fig:dataset}
\end{figure}

\subsubsection{Training the real-time drone detector}\label{sec:training_detector}

We train a single-stage convolutional object detector on the automatically annotated image dataset to obtain drone detections.
We opt for the YOLOv3-tiny architecture \cite{redmon_yolov3_2018} due to its favorable tradeoff between detection accuracy and inference speed on embedded devices.
The network architecture is comprised of a total of $13$ convolutional layers that are interspersed with max-pooling operations and leaky rectified linear units (ReLUs).
The final detections are computed independently at two different scales and subsequently filtered by non-maximum suppression to account for bounding box overlaps.

We make a few notable modifications to the training procedure of the original publication \cite{redmon_yolov3_2018}.
Firstly, we replace the mean-squared error localization loss with an objective that is based on the generalized intersection over union (GIoU) \cite{rezatofighi_generalized_2019} since it provides scale invariance.
Secondly, we employ two recently popularized data augmentation techniques to improve detection accuracy: 1) multi-scale training, and 2) mosaic augmentation \cite{bochkovskiy_yolov4_2020}.
In multi-scale training, each training batch is scaled at random by up to $\pm 50\%$ of its side length to make the detector invariant to object scales.
Mosaic training refers to concatenating four random training samples along their spatial dimensions to obtain an image collage.
The resulting four-image mosaic is subsequently cropped randomly in the center to obtain a new training sample.
We use holdout cross-validation to find suitable values for the most critical hyperparameters.

The parameters of the network are initialized with weights that are pre-trained on the COCO \cite{lin_microsoft_2014} dataset.
The detector is then fine-tuned on a single \tit{drone} class with stochastic gradient descent and Nesterov momentum of $\momentum = 0.937$.
We modulate the learning rate using a cosine annealing schedule with an initial learning rate $\lr = 0.01$ and a final learning rate $\lr_f = 0.0005$.
We also employ a weight decay term of $\wdecay = 0.0005$.
The total loss is computed as

\begin{equation}
    \loss = k^\txt{bal} \loss_{\text{loc}} + \loss_{\text{obj}}
\end{equation}
where $\loss_{\text{loc}}$ denotes the GIoU-based localization loss and $\loss_{\text{obj}}$ refers to the binary cross-entropy objectness loss.
We set the hyperparameter $k^\txt{bal} = 0.055$ to balance the losses and to account for their different magnitudes during training.
Note that unlike the original paper, we omit the classification loss term since we are training on a single class.

Our highest-scoring model achieves an average precision (AP@0.5) of $98.9\%$ \cite{everingham_pascal_2010} at a confidence threshold of $\prob^\txt{conf} = 0.001\%$ on the hold-out test set after $77$ epochs of training.
The model can be trained in less than $\SI{1}{\hour}$ using a batch size of $64$ on a recent GPU such as the Nvidia GeForce RTX 2080Ti.
We also evaluate the model at different image scales on the onboard computer (Sec.~\ref{sec:hardware}) to determine a reasonable speed/accuracy tradeoff of the detector.
We find that performing inference at a resolution of $512 \times 384 $ pixels (in batches of four, one image per camera) provides accurate detections at a frequency of around $\SI{5}{\hertz}$.
For the experiments, we set the confidence threshold to $\prob^\text{conf} = 50\%$ and use a non-maximum suppression threshold of $\prob^\text{nms} = 60\%$.

\subsection{Multi-agent localization and tracking}\label{sec:localization}

\subsubsection{Relative localization based on known physical size}\label{sec:relative_localization}

We compute the relative location of the drone detections using their apparent size in the field of view and the camera parameters.
The camera parameters are obtained using the Kalibr visual-inertial calibration toolbox \cite{furgale_unified_2013}.
We use the equidistant camera model for its compatibility with fisheye lenses and its resulting sub-pixel reprojection errors.

To obtain the relative position estimate from the detection, we first compute the unit-norm bearing vector to the center $\bm{\bearing}^\txt{ctr}$ and one of the extreme points $\bm{\bearing}^\txt{ext}$ of the image bounding box from the intrinsic camera parameters.
We assume the drone can be enclosed by a bounding cube with side length $l$ which is reasonable given its mechanical design (Fig.~\ref{fig:drone_hardware}).
We can then compute the approximate distance to the three-dimensional center of the detected object as

\begin{equation}
    \range = \frac{l / 2}{\tan({\alpha})} + l / 2
\end{equation}
where $\alpha = \cos^{-1}(\bm{\bearing}^\txt{ctr} \cdot \bm{\bearing}^\txt{ext})$ denotes the angle between the unit-norm bearing vectors.
Note that the second term in the above equation accounts for the depth of the object.

\subsubsection{Multi-agent state estimation using random finite sets}\label{sec:multi_agent_tracking}

We use the Gaussian mixture probability hypothesis density (GM-PHD) filter \cite{vo_gaussian_2006} to filter out spurious false-positive detections and to estimate the positions and velocities of nearby agents over time.
We briefly describe the workings of the filter but refer the reader to \cite{vo_gaussian_2006} for more details.
We omit the subscript $i$ to denote the dependence on the focal agent for notational brevity.
The following steps are computed independently for each agent in a decentralized fashion.

\paragraph{Theory}

The GM-PHD filter takes a set of relative localization measurements $\mathcal{Z}_k = \{ \obs_{k,1}, \dots, \obs_{k,M_k} \}$ as input and computes an output set of agent states $\mathcal{X}_k = \{ \state_{k,1}, \dots, \state_{k,J_k} \}$ for each discrete time step $k$.
The states can be described as a single intensity that consists of a weighted sum of Gaussian components of the form
\begin{equation}
    v_k(\state_k) = \sum_{i=1}^{J_k} \weight_k^{(i)} \normal(\state_k; \mean_k^{(i)}, \cov_k^{(i)})
\end{equation}
where $\weight_k^{(i)}$ denotes the weight associated with each of the $J_k$ Gaussian components which are described by their mean $\mean_k^{(i)}$ and covariance $\cov_k^{(i)}$.
Each of the Gaussian components is then propagated with a \tit{prediction} and \tit{update} step, similar to the Kalman filter.

The \tit{prediction} step can be formalized as
\begin{equation}
    v_{k|k-1}(\state_k) = \sum_{i=1}^{J_k} \weight_{k|k-1} \normal(\state_k; \mean_{k|k-1}^{(i)}, \cov_{k|k-1}^{(i)}) + \birth(\obs_k)
\end{equation}
with respective weight, mean, and covariance
\begin{align}
    \weight_{k|k-1}^\ii &= p_{s,k} \weight_{k-1}^\ii \\
    \mean_{k|k-1}^\ii &= \F_{k-1} \mean_{k-1}^\ii + \B_{k-1} \controlinput_{k-1} \\
    \cov_{k|k-1}^\ii &= \F_{k-1} \cov_{k-1}^\ii \F_{k-1}^\top + \Q_{k-1}
\end{align}
where $\F_{k-1}$ is the state transition matrix and $\Q_{k-1}$ the process noise covariance.
To account for the egomotion of the observing drone, we include a control input matrix $\B_{k-1}$ and its control input $\controlinput_{k-1}$.
We let $p_{s,k}$ denote the probability that a Gaussian component survives the prediction step.
We assume an adaptive agent birth model $\gamma_k(\obs_k)$ in which each observation generates a new Gaussian component with weight $\weight_{\birth}$, mean $\mean_{\birth}$, and covariance $\cov_{\birth}$.
We further assume that new agents cannot be spawned from existing ones and that there are no spontaneous births without associated measurement.

The \tit{update} step can be formalized as
\begin{align}
    v_{k}(\state_k) &= (1 - p_{d,k}) v_{k|k-1}(\state_k) \\
    &+ \sum_{\obs_k \in Z_k} \sum_{i=1}^{J_k} \weight_k^\ii (\obs_k) \normal(\state_k; \mean_{k|k}^\ii (\obs_k), \cov_{k|k}^\ii)
\end{align}
with respective weight, mean, and covariance
\begin{align}
    \weight_k^\ii(\obs_k) &=
        \frac   {\prob_{d,k} \weight_{k|k-1}^\ii q_k^\ii (\obs_k)}
                {\clutter_k(\obs_k) + \sum_{j=1}^{J_{k|k-1}} \prob_{d,k}  \weight_{k|k-1}^\jj q_k^\jj(\obs_k)} \\
    \mean_{k|k}^\ii (\obs_k) &= \mean_{k|k-1}^\ii + \K_k^\ii (\obs_k - \HH_k \mean_{k|k-1}^\ii) \\
    \cov_{k|k}^\ii &= (\ones - \K_k^\ii \HH_k) \cov_{k|k-1}^\ii
\end{align}
and
\begin{align}
    q_k^\ii (\obs_k) &= \normal(\obs_k; \HH_k \mean_{k|k-1}^\ii,\HH_k \cov_{k|k-1}^\ii \HH_k^\top + \R_k) \\
    \K_k^\ii &= \cov_{k|k-1}^\ii \HH_k^\top (\HH_k \cov_{k|k-1}^\ii \HH_k^\top + \R_k)^{-1}
\end{align}
where $\HH_k$ is the measurement matrix and $\R_k$ the measurement noise covariance.
We let $p_{d,k}$ denote the probability that an agent is detected during the update step.
We model false positive detections as clutter $\clutter_k$.

Since the number of Gaussian components increases at each filter iteration, the intensity quickly becomes computationally intractable.
Therefore, we prune the components according to the following three conditions to guarantee fast tracking performance.
Firstly, we discard components with a weight of less than the truncation threshold of $T$.
Secondly, we merge components with Mahalanobis distance less than the merging threshold of $U$.
Finally, we retain only the $J_\txt{max}$ components with the largest weights.

\paragraph{Implementation}

We model the state of each agent as $\state_k = \left[ \pos_{x,k}, \pos_{y,k}, \vel_{x,k}, \vel_{y,k} \right]^\top $ which consists of relative position $(\pos_{x,k}, \pos_{y,k})$ and velocity $(\vel_{x,k}, \vel_{y,k})$.
The position and velocity components of the state are two-dimensional since the agents are assumed to fly in a planar configuration at approximately the same altitude.
The control input $\controlinput_k$ is defined as the linear velocity of the drone in the body frame which we obtain from the internal state estimate of the autopilot.
Finally, the measurements $\obs_k = [ \range_k, \bearing_k ]^\top$ consist of range and bearing, respectively.

The process and measurement noise covariances are modeled as

\begin{align} \label{eq:process_obs_noise}
    \Q_k &= \std_\process^2
        \begin{bmatrix}
            \frac{\dt_k^4}{4} \ones_2 & \frac{\dt_k^3}{2} \ones_2 \\
            \frac{\dt_k^3}{2} \ones_2 & \dt_k^2 \ones_2
        \end{bmatrix} &
        &\txt{and} &
    \R_k &=
        \begin{bmatrix}
            \std_\range^2 & 0 \\
            0 & \std_\bearing^2
        \end{bmatrix}
\end{align}
where $\dt_k$ denotes the time elapsed since the last measurement and is computed as the difference between consecutive timestamps $\dt_k = t_k - t_{k-1}$.
We further let $\std_\process$, $\std_\range$ and $\std_\bearing$ denote the standard deviation of the process, range, and bearing noise, respectively.

The state transition function follows a linear Gaussian model and is defined as

\begin{equation}
    f(\state_{k-1}, \controlinput_k) = \F_k \state_{k-1} + \B_k \controlinput_k
\end{equation}
where
\begin{align}
    \F_k &=
        \begin{bmatrix}
            \ones_2 & \dt_k \ones_2 \\
            \zeros_2 & \ones_2
        \end{bmatrix} &
        &\txt{and} &
    \B_k &= \dt_k \ones_2.
\end{align}
The observation model is nonlinear and consists of measurements of range and bearing

\begin{equation}
    h(\state_k) =
        \begin{bmatrix}
            \range_k \\
            \bearing_k
        \end{bmatrix} =
        \begin{bmatrix}
            \sqrt{\pos_{x,k}^2 + \pos_{y,k}^2} \\
            \atantwo(\pos_{y,k}, \pos_{x,k})
        \end{bmatrix}
\end{equation}
where we use the two-argument function $\atantwo$ to avoid ambiguities in the conversion from cartesian to polar coordinates.
We linearize the measurement model by computing the Jacobian with respect to the state as

\begin{equation}
    \HH_k = \frac{\partial h(\state_k)}{\partial \state_k} =
        \begin{bmatrix}
            \begin{matrix}
                \frac{\pos_{x,k}}{\sqrt{\pos_{x,k}^2 + \pos_{y,k}^2}} & \frac{\pos_{y,k}}{\sqrt{\pos_{x,k}^2 + \pos_{y,k}^2}} \\
                -\frac{\pos_{x,k}}{\pos_{x,k}^2 + \pos_{y,k}^2} & \frac{\pos_{y,k}}{\pos_{x,k}^2 + \pos_{y,k}^2}
            \end{matrix}
            & \zeros_2
        \end{bmatrix}.
\end{equation}

We set the probability of detection to $\prob_d = 90\%$ to provide a slightly more conservative estimate of the detection performance during real-time inference than the results on our test dataset suggest (Sec.~\ref{sec:training_detector}).
We assume a probability of survival of $\prob_s = 100\%$ since agents that are detected once should not disappear.
New Gaussian components are initialized directly from the measurements using an adaptive birth model with weight, mean, and covariance
\begin{align}
    \weight_\birth &= 10^{-5} \\
    \mean_\birth &= \left[ \range_k \cos(\bearing_k), \range_k \sin(\bearing_k), 0, 0 \right]^\top \\
    \cov_\birth &= \diag\left( \left[ \std_\pos^2, \std_\pos^2, \std_\vel^2, \std_\vel^2 \right] \right)
\end{align}
where $\std_\pos$ and $\std_\vel$ are the standard deviation of the position and velocity which we set to $\SI{1}{\meter}$ and $\SI{10}{\meter\per\second}$, respectively.
The mean is computed by converting the raw measurement from polar to cartesian coordinates, assuming zero initial velocity.
We model false positive detections by assuming that we observe one clutter return per time step and therefore set $\clutter_k = 1 / a^2$ where $a = \SI{10}{\meter}$ is the side length of the virtual arena.
Although the bearing measurements are precise to a single degree, we set its standard deviation to a slightly larger value of $\std_\bearing = 3^\circ$ to account for factors such as calibration inaccuracies or imprecisions in the detections due to background clutter.
Since we experimentally determined that the range noise varies with the distance to the observer, we model its standard deviation as a function of the measured distance itself (Sec.~\ref{sec:relative_localization_results}).
Finally, we model the truncation threshold as $T = 10^{-5}$, the merging threshold as $U = 0.5$, and the maximum number of components to $J_\txt{max} = 100$.

\subsection{Flocking algorithm}\label{sec:control}

The method described so far could be leveraged by any flocking algorithm.
In this work, we use a control algorithm based on the Reynolds flocking rules \cite{reynolds_flocks_1987} to compute high-level velocity commands from the relative position estimates of nearby drones \cite{schilling_learning_2019}.
The weighted velocity commands are: 1) a repulsive \tit{separation} term to steer nearby drones away from each other, 2) a \tit{cohesion} term to keep the drones close to each other, and 3) a \tit{migration} term that provides a navigation goal to the swarm.
The respective velocity terms for separation, cohesion, and migration can be formalized as
\begin{align}
    \velvec_i^\txt{sep} &= k^\txt{sep} \frac{1}{|\neighbors_i|} \sum_{j \in \neighbors_i} \frac{-\relvec_{ij}}{\norm{\relvec_{ij}}} \\
    \velvec_i^\txt{coh} &= k^\txt{coh} \frac{1}{|\neighbors_i|} \sum_{j \in \neighbors_i} {\relvec_{ij}} \\
    \velvec_i^\txt{mig} &= k^\txt{mig} \frac{\relvec^\txt{mig}}{\norm{\relvec^\txt{mig}}}
\end{align}
where $\relvec_{ij}$ denotes the relative position of the $j$-th agent with respect to agent $i$, and $\neighbors_i$ the set of neighbors of the $i$-th agent.
The migration point is denoted by $\relvec^\txt{mig}$ and expressed relative to the body frame.
We further let $k^\txt{sep}$, $k^\txt{coh}$, and $k^\txt{mig}$ denote the gains that modulate the strength of the separation, cohesion, and migration terms, respectively.
The final velocity command is then given by $\velvec_i = \tilde{\velvec}_i / \norm{\tilde{\velvec}_i} \min(\norm{\tilde{\velvec}_i}, \vel^\txt{max})$ where $\tilde{\velvec}_i = \velvec_i^\txt{sep} + \velvec_i^\txt{coh} + \velvec_i^\txt{mig}$ is the sum of all velocity terms and $\vel^\txt{max}$ is the desired maximum speed cutoff.

During the experiments, we set the maximum speed to $\vel^\txt{max} = \SI{0.5}{\meter \per \second}$, and the separation, cohesion, and migration gains to $k^\txt{sep} = 7$, $k^\txt{coh} = 1$, and $k^\txt{mig} = 1$, respectively.
The gains are chosen such that the agents converge to an equilibrium distance of approximately $\SI{2}{\meter}$ during migration.

\begin{figure}[t]
    \centering
    \includegraphics[width=\columnwidth]{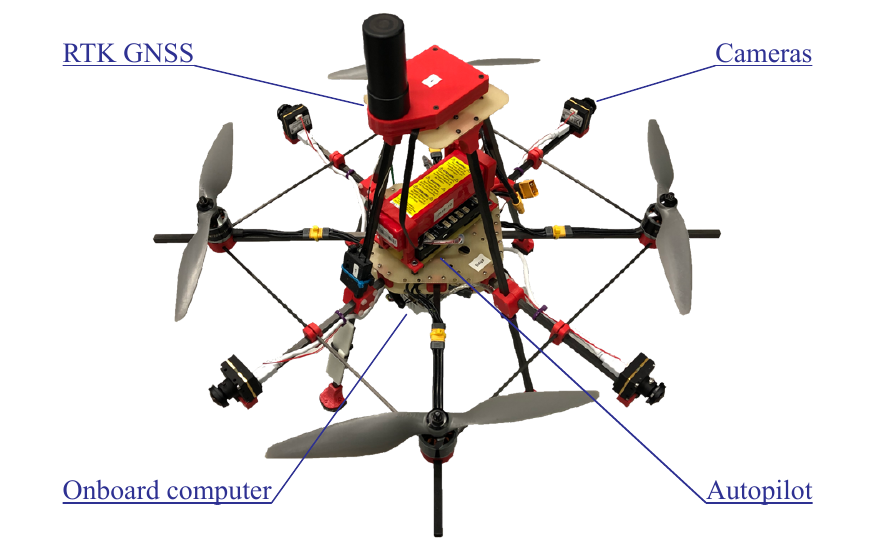}
    \caption{The LeQuad drone features omnidirectional vision via four cameras, a high performance onboard computer with embedded GPU for real-time inference, and a RTK-enabled GNSS to obtain centimeter-accurate ground-truth positions.}\label{fig:drone_hardware}
\end{figure}


\section{Experimental setup}\label{sec:experimental_setup}

\subsection{Hardware}\label{sec:hardware}

We use a custom-built quadcopter named LeQuad for all experiments (Fig.~\ref{fig:drone_hardware}).
Each quadcopter features four FLIR Firefly S global-shutter cameras mounted at a right angle from each other to obtain omnidirectional visual inputs.
Each camera is equipped with an OpenMV ultra-wide angle lens which provides a horizontal and vertical field of view of $ \SI{166}{\degree} $ and $\SI{116}{\degree}$, respectively.
We operate the cameras over a powered USB 2 hub at a binned resolution of $720 \times 540$ to obtain grayscale images at a frequency of $\SI{10}{\hertz}$.
We refrain from using USB 3 to avoid electromagnetic interference with the Drotek F9P RTK-GNSS receiver, which provides centimeter-accurate absolute positions.
We use the Nvidia Jetson TX2 mounted on a ConnectTech Orbitty carrier board as an onboard computer and the Holybro Pixhawk 4 as an autopilot.

\subsection{Software}\label{sec:software}

The onboard computer runs Ubuntu 18.04 bundled with the Linux4Tegra (L4T) distribution, and we use ROS Melodic as a robotics middleware.
The autopilot runs PX4 and is responsible for hardware-triggering the cameras to provide the resulting images with IMU-synchronized timestamps.
The neural networks are trained and evaluated using PyTorch.


\section{Results}\label{sec:results}

We first report the accuracy of the visual relative localization of drones in a controlled indoor environment (Sec.~\ref{sec:relative_localization_results}) and then show vision-based flocking with three real quadcopters performing several navigation tasks outdoors (Sec.~\ref{sec:outdoor_results}).

The proposed approach has also been extensively tested using the Gazebo simulator with up to ten vision-based agents.
For instance, we simulated noisy detections with different false positive and negative rates, as well as processing delays, to find suitable parameter values and operational requirements for the detection, tracking, and control modules.
However, we find that it is difficult to realistically model misdetections in simulation since their distribution highly depends on environmental conditions such as visual clutter.

\subsection{Visual relative localization}\label{sec:relative_localization_results}

We show results on the theoretical performance of the visual relative localization system to test its operational bounds and to find suitable values for the range and bearing noise parameters $\std_{\range}$ and $\std_{\bearing}$ (Eq.~\ref{eq:process_obs_noise}).
To this end, we employ a setup similar to the one used for automatic labeling (Sec.~\ref{sec:background_subtraction}) except that we additionally obtain ground truth poses of the observing camera and drone from a motion capture system.
After transforming the visual detections into the frame of the motion capture system, we can directly compare the drone's true position with its estimate obtained using vision in metric space.
We find that the relative localization error varies considerably as a function of the distance to the drone, whereas the error caused by bearing variations is negligible (Fig.~\ref{sec:relative_localization}).

\newcommand{\locwidth}{0.48 \columnwidth}
\begin{figure}[t]
    \begin{subfigure}{\locwidth}
        \includegraphics[width=\textwidth]{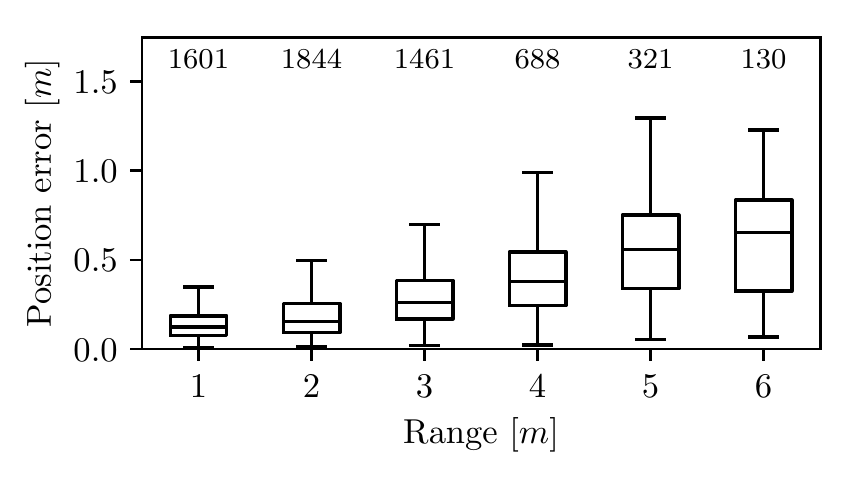}
    \end{subfigure}
    \hfill
    \begin{subfigure}{\locwidth}
        \includegraphics[width=\textwidth]{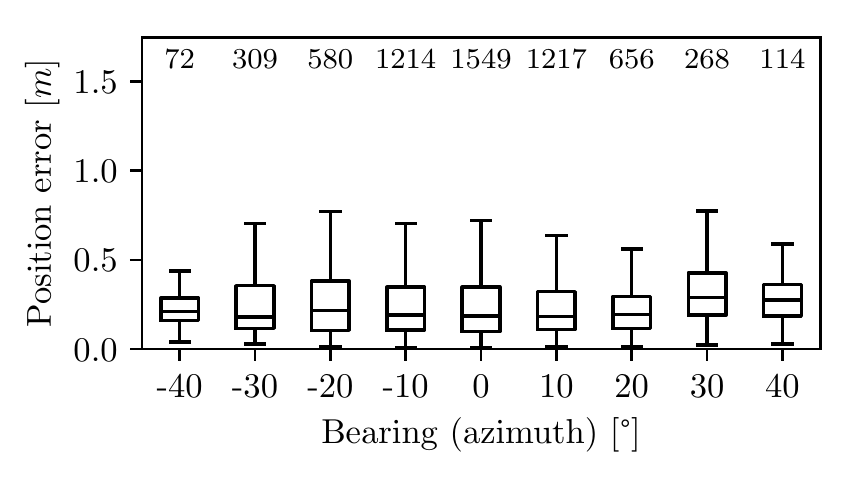}
    \end{subfigure}
    \caption{%
        Comparison of vision-based relative localization errors by \tit{range} (left) and \tit{bearing} (right).
        The bearing estimates are near-constant over the field of view, whereas the range errors increase with distance from the observer.
        Millimeter-accurate ground-truth positions are obtained at $\SI{100}{\hertz}$ using a motion capture system.
        The counts above the boxes indicate the number of measurements used to calculate their statitics.
    }\label{fig:localization}
\end{figure}

\subsection{Collective outdoor navigation}\label{sec:outdoor_results}

\newcommand{\fwidth}{0.32 \textwidth}
\begin{figure*}[t]
    \begin{subfigure}{\fwidth}
        \caption{}\label{fig:migration_linear_trajectories}
    \end{subfigure}
    \begin{subfigure}{\fwidth}
        \caption{}\label{fig:migration_rectangular_trajectories}
    \end{subfigure}
    \begin{subfigure}{\fwidth}
        \caption{}\label{fig:migration_circular_trajectories}
    \end{subfigure}
    \\
    \begin{subfigure}{\fwidth}
        \includegraphics[width=\textwidth]{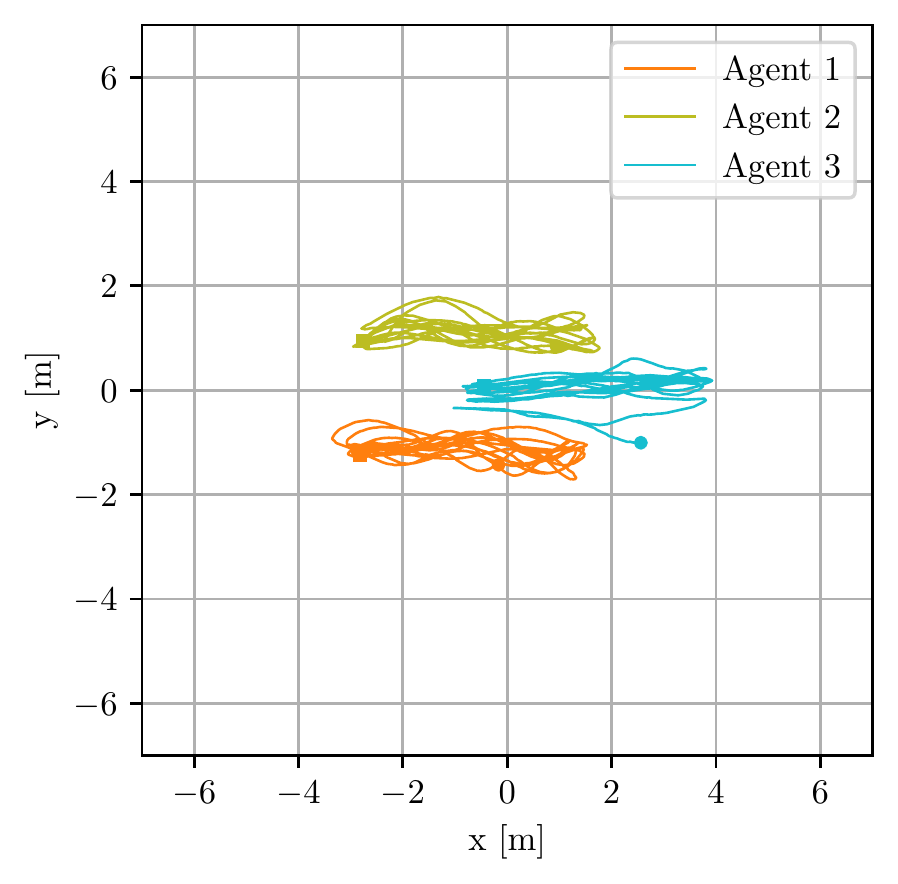}
    \end{subfigure}
    \hfill
    \begin{subfigure}{\fwidth}
        \includegraphics[width=\textwidth]{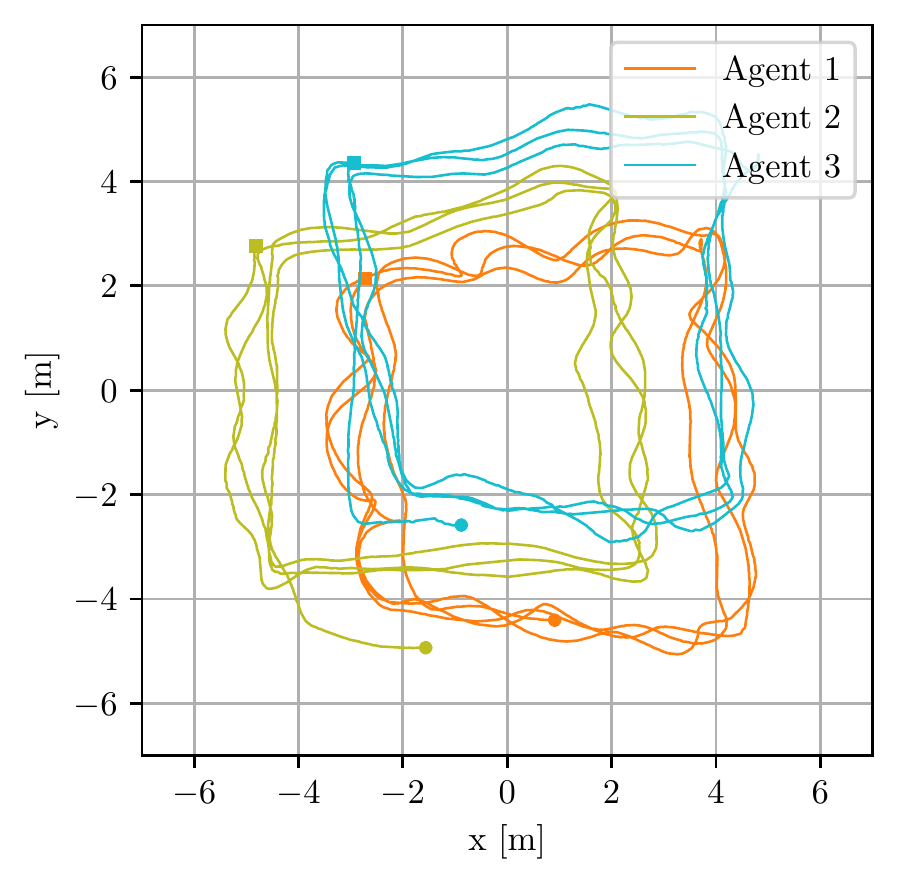}
    \end{subfigure}
    \hfill
    \begin{subfigure}{\fwidth}
        \includegraphics[width=\textwidth]{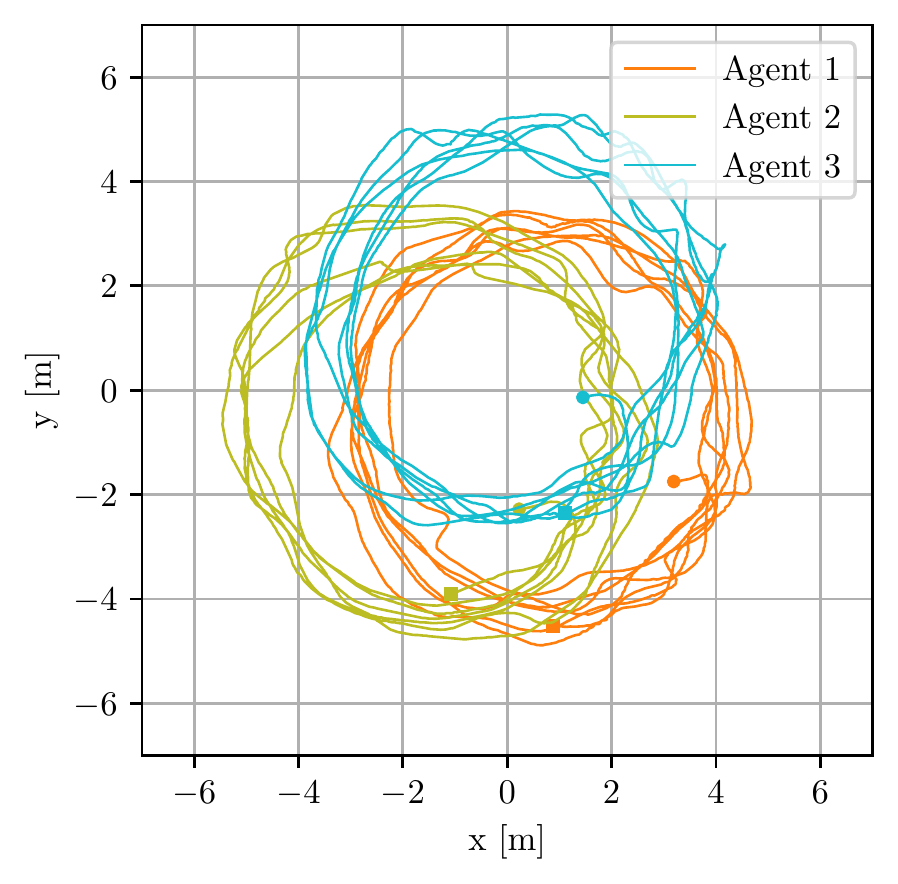}
    \end{subfigure}
    \\
    \begin{subfigure}{\fwidth}
        \includegraphics[width=\textwidth]{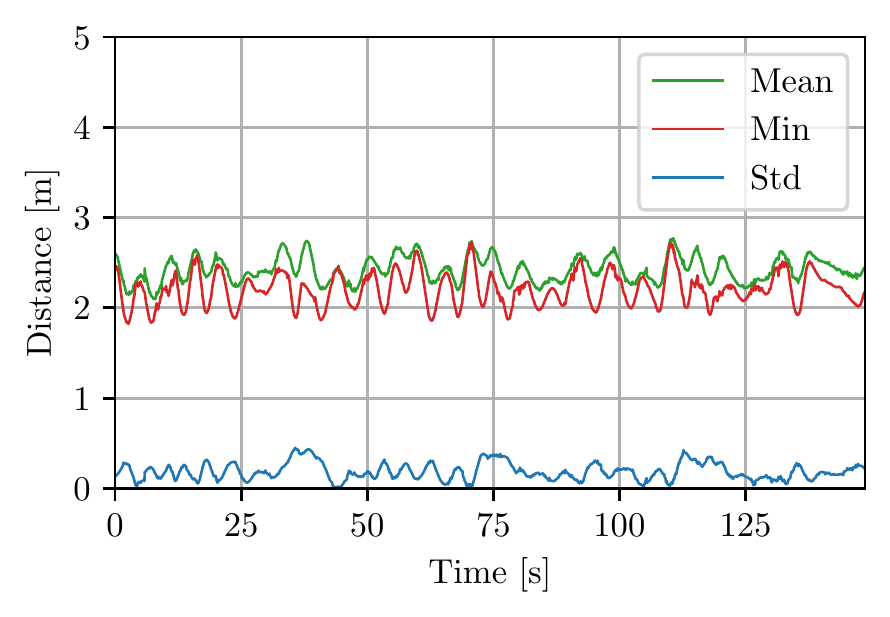}
    \end{subfigure}
    \hfill
    \begin{subfigure}{\fwidth}
        \includegraphics[width=\textwidth]{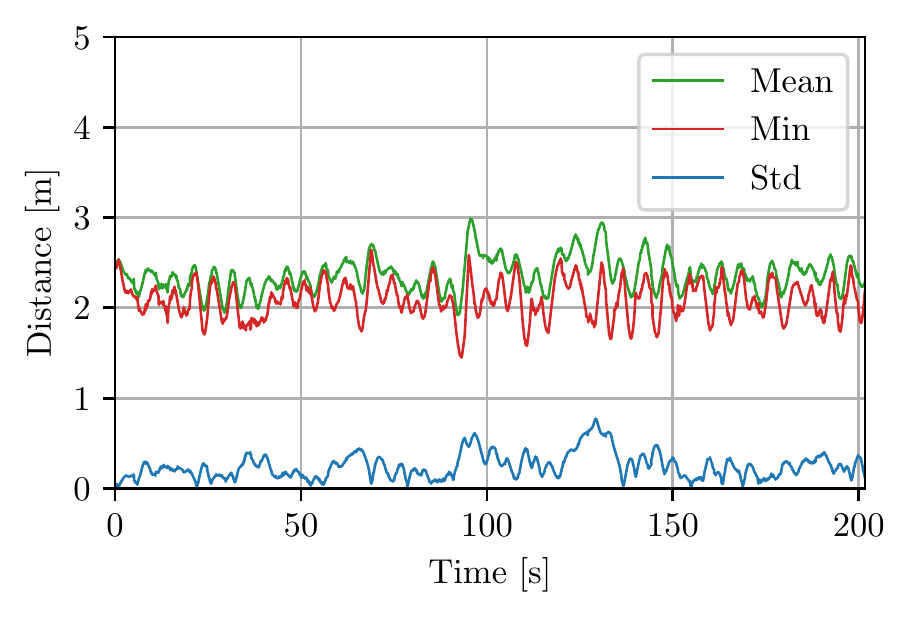}
    \end{subfigure}
    \hfill
    \begin{subfigure}{\fwidth}
        \includegraphics[width=\textwidth]{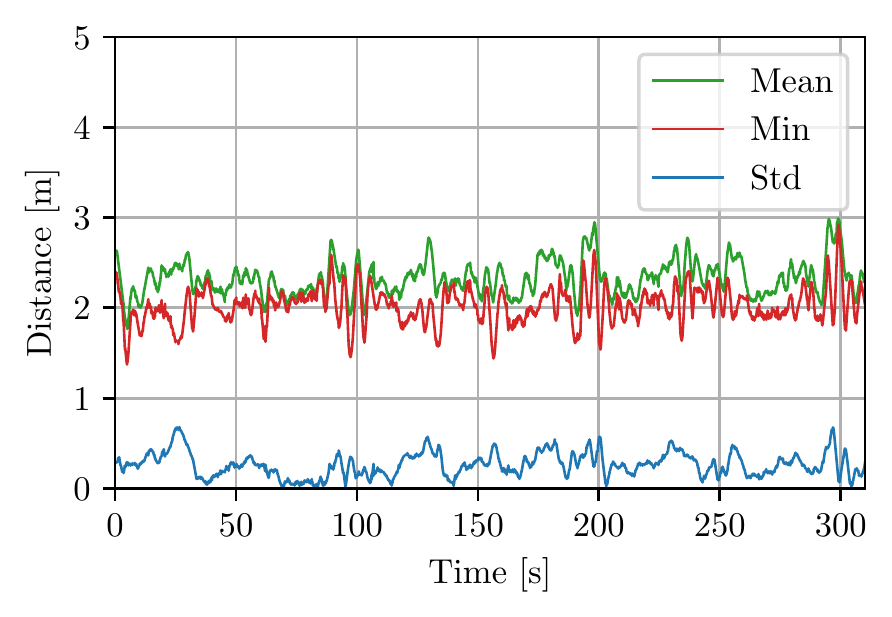}
    \end{subfigure}
    \\
    \begin{subfigure}{\fwidth}
        \caption{}\label{fig:migration_linear_distances}
    \end{subfigure}
    \begin{subfigure}{\fwidth}
        \caption{}\label{fig:migration_rectangular_distances}
    \end{subfigure}
    \begin{subfigure}{\fwidth}
        \caption{}\label{fig:migration_circular_distances}
    \end{subfigure}
    \caption{%
        Results of vision-based flocking in outdoor environments.
        We show \tit{trajectories} (top) and inter-agent \tit{distances} (bottom) for three different migration scenarios: \tit{linear} (left), \tit{rectangular} (middle) and \tit{circular} (right) migration.
        The drones remain collision-free and cohesive using only local visual information, which is processed onboard in real-time.
        Centimeter-accurate ground-truth positions are obtained at $\SI{10}{\hertz}$ using RTK-enabled GNSS receivers mounted on the drones.
        These positions are not shared between the drones during the experiments and only serve to analyze the inter-agent distances.
    }\label{fig:migration}
\end{figure*}

\newcommand{\dwidth}{0.24 \textwidth}
\begin{figure*}[t]
    \begin{subfigure}{\dwidth}
        \includegraphics[width=\textwidth]{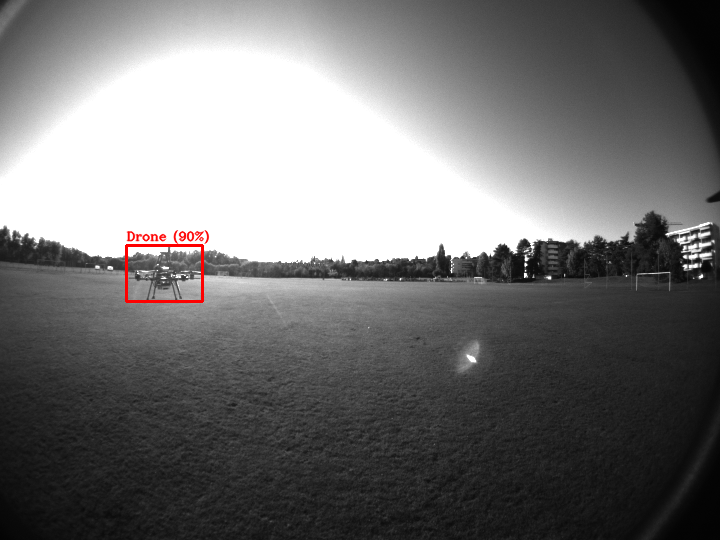}
    \end{subfigure}
    \hfill
    \begin{subfigure}{\dwidth}
        \includegraphics[width=\textwidth]{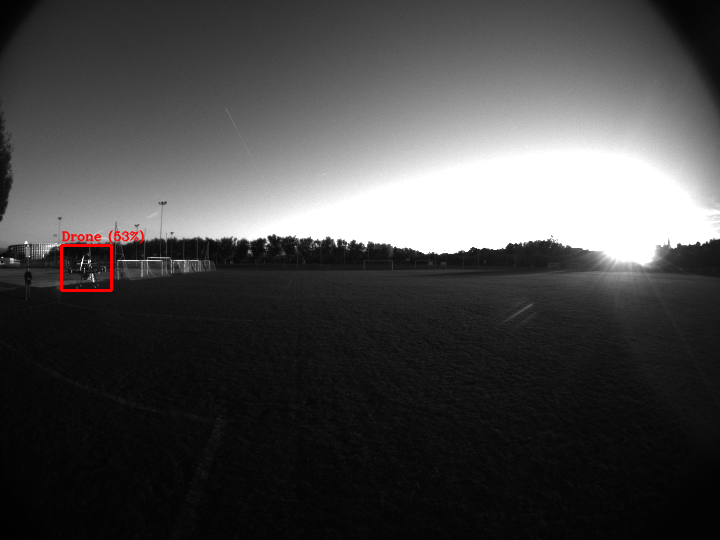}
    \end{subfigure}
    \hfill
    \begin{subfigure}{\dwidth}
        \includegraphics[width=\textwidth]{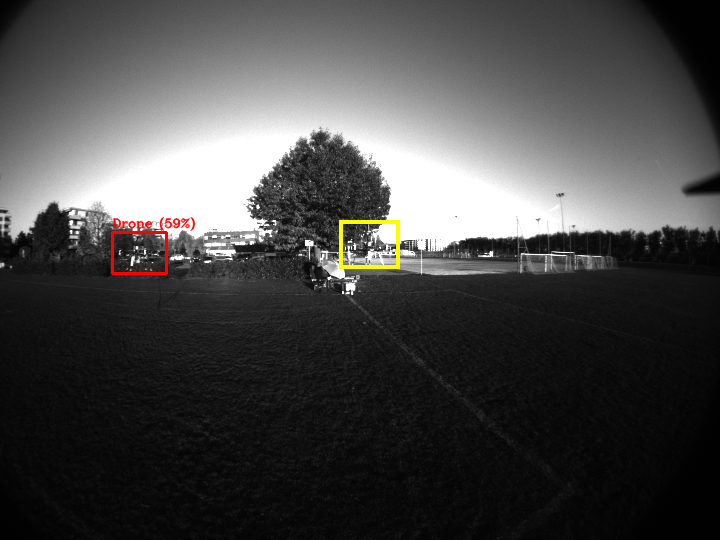}
    \end{subfigure}
    \hfill
    \begin{subfigure}{\dwidth}
        \includegraphics[width=\textwidth]{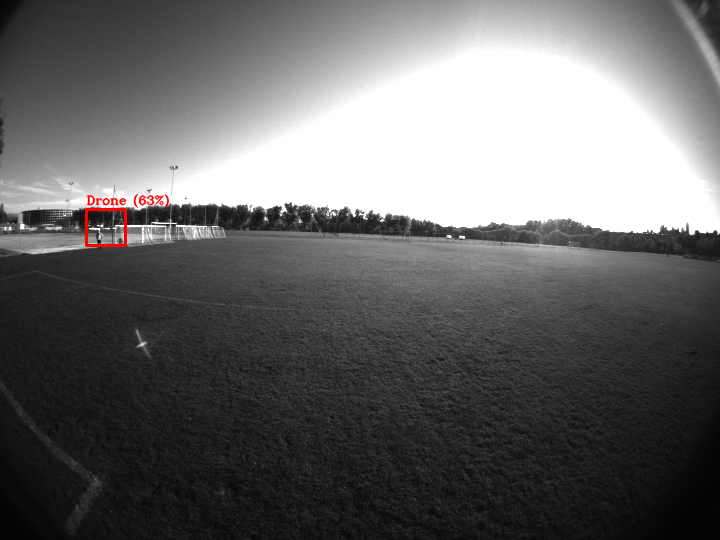}
    \end{subfigure}
    \\
    \begin{subfigure}{\dwidth}
        \caption{High confidence}\label{fig:detection_easy}
    \end{subfigure}
    \begin{subfigure}{\dwidth}
        \caption{Low confidence}\label{fig:detection_difficult}
    \end{subfigure}
    \begin{subfigure}{\dwidth}
        \caption{False negative}\label{fig:detection_false_negative}
    \end{subfigure}
    \begin{subfigure}{\dwidth}
        \caption{False positive}\label{fig:detection_false_positive}
    \end{subfigure}
    \caption{%
        Qualitative examples of detection results categorized by confidence score (high and low) and type of error (false negative and false positive).
        High confidence (\ref{fig:detection_easy}): the drone is easily detected in front of the grass texture despite direct sunlight.
        Low confidence (\ref{fig:detection_difficult}): the drone blends in with the background due to the line features and lighting conditions.
        False negative (\ref{fig:detection_false_negative}): the right drone (yellow, manually labeled) can not be reliably distinguished from the background clutter.
        False positive (\ref{fig:detection_false_positive}): this type of spurious detection occurs relatively frequently but is filtered out by the multi-agent state tracker.
    }\label{fig:detections}
\end{figure*}

We report results for three different navigation scenarios: \tit{linear}, \tit{rectangular}, and \tit{circular} migration.
Before each flight, we place the drones at roughly $\SI{2.5}{\meter}$ distance from each other and wait for their RTK-GNSS receivers to converge to a fixed solution which provides centimeter-accurate absolute positions at $\SI{10}{\hertz}$.
These measurements are only used to provide a reliable ground-truth for the evaluation of the experiments.

After the RTK fix is obtained, we let all agents take off simultaneously and reach an altitude of $\SI{2}{\meter}$ before we let the vision-based flocking algorithm take over the control of their motion.
The agents are given the same list of migration points depending on the type of navigation scenario.
We switch from one migration point to the next as soon as an agent enters an acceptance radius of $r^\txt{acc} = \SI{3}{\meter}$.
As the list of migration points is exhausted, we repeat the procedure from the first waypoint.
We stop the experiment as soon as the battery level of one of the agents reaches a critical capacity of $15\%$.

The agents' altitude is individually regulated using a proportional controller to constrain their motion to a horizontal plane.
However, the planar constraint may be lifted by mounting additional cameras that point to the top and bottom, or by equipping the existing cameras with lenses that provide a larger field of view.

During the linear migration experiment, the agents fly between two waypoints that are located $\SI{10}{\meter}$ apart from each other (Fig.~\ref{fig:migration_linear_trajectories}).
Over a total flight duration of around $\SI{2.5}{\minute}$, the minimum inter-agent distance the agents reach is $\SI{1.82}{\meter}$ and the overall mean $\SI{2.42}{\meter}$ (Fig.~\ref{fig:migration_linear_distances}).

The rectangular migration experiment defines four waypoints that are located at the corners of a square with side length $\SI{10}{\meter}$ (Fig.~\ref{fig:migration_rectangular_trajectories}).
The total flight time is around $\SI{3.3}{\minute}$ and the overall minimum and mean inter-agent distances are $\SI{1.45}{\meter}$ and $\SI{2.36}{\meter}$, respectively (Fig.~\ref{fig:migration_rectangular_distances}).

Finally, the circular migration experiment leads the agents through a series of twelve waypoints that are linearly spaced around a circle with $\SI{10}{\meter}$ diameter (Fig.~\ref{fig:migration_circular_trajectories}).
During an overall flight time of around ${\SI{5}{\minute}}$, the agents remain collision-free while reaching a minimum inter-agent distance of $\SI{1.37}{\meter}$ and an overall mean distance of $\SI{2.32}{\meter}$ (Fig.~\ref{fig:migration_circular_distances}).

The above experiments are three representative flights taken from a total of $\SI{30}{\min}$ of collision-free experimental recordings.
The progressively lower inter-agent distances across linear, rectangular, and circular migration experiments can be explained by examining the distribution and/or density of waypoints.
During the rectangular migration experiment, the lowest inter-agent distances are reached close to the corners where directional changes occur.
In the case of the circular migration experiment, the larger number and density of waypoints have a cohesive effect since the agents simultaneously approach points that are more densely spaced.

During the experiments, a variety of objects that can potentially confuse the detector are present in the background (Fig.~\ref{fig:detections}).
In descending order of frequency, the most common objects aside from drones are trees, buildings, cars, people, fences, traffic signs, tables, and even dogs.
The other drones are generally detected despite background clutter and adverse lighting conditions (Fig.~\ref{fig:detection_easy} and \ref{fig:detection_difficult}).
False-negative detections occur most frequently when another drone is flying far away and/or in front of the ground control station, i.e. a camping table with the experimental equipment covered by a sun umbrella (Fig.~\ref{fig:detection_false_negative}).
False-positive detections appear most often in image regions that contain many line-like features since they resemble the mechanical design of the drone (Fig.~\ref{fig:detection_false_positive}).
They are mostly present for the duration of a single frame and the tracker can reliably reject them.
False tracks are occasionally created if false-positive detections are present for more than one frame.
However, the false tracks do not cause instabilities that lead to collisions.


\section{Conclusions}\label{sec:conclusions}

We presented a vision-based detection and tracking algorithm that enables dense groups of drones to fly cohesively and without mutual collisions.
The proposed approach does not depend on visual markers or inter-agent communication and is thus suitable for flocking operation in GNSS-denied environments or in situations where wireless links are unreliable.
The approach is fully decentralized since each agent relies exclusively on onboard processing of local visual information to estimate the positions and velocities of neighboring drones.
The outdoor navigation experiments show that the system is robust to background clutter and enables collision-free flight even in demanding lighting conditions.
These experiments should be considered as minimal validation conditions of vision-based flocking algorithms without explicit communication or localization infrastructure.
Further experiments are needed to ensure the scalability of the proposed system to larger numbers of vision-based agents.


\section*{Acknowledgments}\label{sec:acknowledgments}

We thank Olexandr Gudozhnik for the contributions to the drone hardware, Vivek Ramachandran and Enrico Ajanic for the help with conducting outdoor experiments, as well as Enrica Soria, Valentin Wüest, and Davide Zambrano for the helpful discussions.

\balance

\bibliographystyle{IEEEtran}
\bibliography{library/strings/conferences-abrv,library/strings/journals-abrv,library/library}

\end{document}